\documentclass{llncs}

\usepackage{color}

\begin{document}

\title{ The Temporal Singularity: time-accelerated simulated civilizations and their implications }
\titlerunning{ The Temporal Singularity }  
%


\author{Giacomo Spigler\inst{1} }
\authorrunning{Giacomo Spigler} 
\tocauthor{Giacomo Spigler}
\institute{The Biorobotics Institute, Scuola Superiore Sant'Anna, Pisa, Italy,\\
\url{http://www.spigler.net/giacomo}, \\
\email{giacomo.spigler@santannapisa.it} }

\maketitle              

\begin{abstract}

Provided significant future progress in artificial intelligence and computing, it may ultimately be possible to create multiple Artificial General Intelligences (AGIs), and possibly entire societies living within simulated environments. In that case, it should be possible to improve the problem solving capabilities of the system by increasing the speed of the simulation. If a minimal simulation with sufficient capabilities is created, it might manage to increase its own speed by accelerating progress in science and technology, in a way similar to the Technological Singularity. This may ultimately lead to large simulated civilizations unfolding at extreme temporal speedups, achieving what from the outside would look like a Temporal Singularity. Here we discuss the feasibility of the minimal simulation and the potential advantages, dangers, and connection to the Fermi paradox of the Temporal Singularity. The medium-term importance of the topic derives from the amount of computational power required to start the process, which could be available within the next decades, making the Temporal Singularity theoretically possible before the end of the century.

\end{abstract}


\section{The Temporal Singularity}

It seems possible, if not likely, that artificial agents with general intelligence (AGI) will be built in the future \cite{kurzweil2005singularity,muller2016future}. It also seems likely that such agents could be further improved to achieve super-human degrees of intelligence (ASI).  A simple way to increase the capabilities of an agent is to execute the same algorithms on a faster (super-)computer, so to provide it with more time to think and solve problems, thus resulting in a shorter solving time in the external world. In practice, a simulated environment may be required for the agent to work in, as the `slow' external world would be a limitation to the performance the agent even at moderate speedups. It is interesting to note that this approach is already regularly used, for example in training deep reinforcement learning (DRL) agents, that is usually performed in simulated environments whose execution speed is limited only by the available computing power \cite{openai_gym,deepmind_lab,microsoft_malmo}. For example, DRL agents learning to play Atari games can experience thousands of game frames per second even on a regular desktop computer, compared to human players that play them at 15-60 frames per second.


Another approach to improve the effective capabilities of the system without any modification to its algorithms is to simulate multiple agents each with its specific differences, so that they can come up with different ways of solving the problem individually or cooperatively by exploiting dynamics of collective intelligence. An interesting outcome of simulations of this type is the potential to simulate the unfolding of entire ``civilizations'', possibly pursuing complex sets of goals like general progress in science and technology. The potential of the approach relies not only on the possibly advanced intelligence level of the agents (ASI), but also on the temporal speedups that could be achieved by increasing the computing resources available for the simulation. Throughout this paper intelligent agents and civilizations will be referred to as `simulated' only to mean that they experience a simulated environment in contrast to the `real' external world, but there is no reason not to consider them as real as any intelligent agent outside the simulation.

Here we suggest that \emph{if} it will be possible to create at least a limited group of AGIs in a simulation unfolding faster than the external time, \emph{then} such simulation may be able to accelerate the rate of progress in science and technology, possibly by continually self-improving its core technologies such as its intelligence algorithms and its computing systems in a manner similar to the \emph{Technological Singularity} \cite{good1966speculations,vinge1993coming,kurzweil2005singularity,hutter2012can,chalmers2010singularity}, and thus potentially achieve a runaway increase in its capabilities. Specifically, the rate of progress may be so high that in a very short time the simulations could progress to producing entire civilizations spanning thousands or millions of years or even more in an arbitrarily short time interval elapsed in the external world, achieving what from the outside would be a \textbf{Temporal Singularity}. In particular the Temporal Singularity is defined as the moment in time where a minimal simulation capable of beginning the runaway exponential self-improvement is started. We will discuss the feasibility of the minimal simulation in Section \ref{sec:feasibility}.

It is difficult to imagine what such a quick progress would look like, as even a single century of progress at the present rate is challenging to forecast. Even more, we can only wonder what the world would become after the Temporal Singularity has allowed the unfolding of millions or billions of years of an advanced civilization \cite{hutter2012can}, during which potentially any questions our species may ever ask could have been answered. 

This result is compatible with the idea of the Technological Singularity, of which the Temporal Singularity can represent a component or a way to achieve it. Contrary to the main definitions of the Technological Singularity, however, the Temporal Singularity would not necessarily require a runaway increase in the cognitive capabilities of the artificial agents, but rather only a runaway increase in the temporal speedups of the simulations. We should note that speeding up the execution of AGIs has been already suggested in this context, for example by Vernor Vinge, who discusses an AI whose `mind clock' is significantly faster than its creator and the problem of AI boxing \cite{vinge1993coming}, or by Solomonoff in the context of an exponential increase in the number of simulated agents \cite{solomonoff1985time}. Most notably Marcus Hutter \cite{hutter2012can} explored what the Technological Singularity would look like for both the outside and the inside of a virtual software society undergoing it, also discussing the difference between speeding up the simulation time and increasing the intelligence of the agents. However, the focus of the discussion was put on the extreme progress and changes achieved in the traditional Technological Singularity, rather than on the implications of a drastic increase in the temporal speedups of the simulations and its potential implications on the Fermi Paradox. 

The idea of simulated civilizations is also not novel, although it has been generally applied to us being in the simulation ourselves, rather than focusing directly on the benefits, limits and implications of us producing it, and in particular on the possibility to speed up the elapsing of the simulated time. Philosophers have always wondered about the nature of reality and the possibility of it being an illusion. In recent times, the argument has been especially developed by Hans Moravec \cite{moravec1999simulation} and Nick Bostrom \cite{bostrom2003we} in the explicit context of computer simulations. A more closely related investigation was proposed by Vidal, who explored the possibility that scientific simulations will improve significantly in the future and finally result in simulating an entire universe, in order to better probe and understand our own universe and the processes of physical, biological and cultural evolution \cite{vidal2008future}. However, most of the discussions such as Vidal's and Bostrom's only focus on a very special type of simulations restricted to detailed versions of our physical universe and our same society and life as we know it, which although intriguing from a scientific point of view, constitute only a tiny fraction of the potential uses of time-accelerated simulations, and possibly an inefficient use of the computing resources. For example, as we discuss in section \ref{sec:feasibility}, it may be that fooling the simulated agents to prevent them from discovering that they belong to a simulation may not be necessary, which would in turn lower the computational requirements for the simulated environment. In any case, whether our own world is itself simulated or not does not reduce the potential advantages of running our own time-accelerated simulations.

Section \ref{sec:feasibility} will next overview the feasibility and broad computational requirements for simulations capable of achieving and sustaining the Temporal Singularity, while Section \ref{sec:implications} will explore some of the advantages and risks of such simulations, and the implications of the Temporal Singularity for the Fermi paradox.

\section{Feasibility}
\label{sec:feasibility}

\textbf{The minimal simulation. } It is difficult to estimate what are the minimal requirements for a simulation capable of starting the runaway exponential process of self-improvement and time-acceleration that characterizes the Temporal Singularity. In general, we should expect the minimal simulation to provide a problem-solving capability sufficient to compete with teams of human experts, either by providing significant temporal speedups, by using more capable AGIs/ASIs or by creating a larger number of individuals. Even small advantages, compared to traditional research and development, may be sufficient to start the process by exploiting the compound nature of progress \cite{kurzweil2005singularity,solomonoff1985time}. The minimal requirements could thus be reasonably low (see the discussion on the \textit{computational requirements} below), especially after achieving human-level AGI, which itself however \emph{may} not be required, as a super-human narrow intelligence in specific fields like improving the computing technology may be sufficient.

\textbf{AGI. } Still, while we could imagine some limited type of ``civilization'' composed by agents with narrow intelligence (ANI), the development of artificial general intelligence (AGI) is likely to be a core requirement for enabling complex artificial civilizations. It is not known whether AGI itself will ever be possible, though there do \emph{not} seem to be strong reasons for it to be not. Unfortunately, the field is known to have a poor track record of predictions about when such a system wil be developed. Current predictions also vary greatly depending on the expected requirements for specific types of implementations, with average agreement placed around 2040 \cite{muller2016future,bostrom2014superintelligence} and possibly as early as 2029 \cite{kurzweil2005singularity,barrat2013our}, and a high confidence in any case that it may happen before the end of the century. We could also wonder whether the artificial agents could instantiate \emph{consciousness}, but it may not be a strict requirement in this context. On the other hand, it may turn out that consciousness is required, for example for the establishment and maintenance of societies and complex civilizations (e.g., for consciousness and sociality \cite{graziano2013consciousness}). 



\textbf{Fooling the agents. } The requirements for the simulations discussed here also change significantly depending on whether the simulated agents are allowed to know they belong to a simulation or whether they need to be fooled. In particular, fooling the agents may be challenging especially if the aim of the simulations is to produce progress in science and technology that apply to the external world, as a large degree of knowledge of it would be required. In the limit, a perfect simulation of our physical world may be required for perfect fooling, which would however limit the simulation (for an analysis of the requirements, see for example \cite{baxter2001planetarium}). It is however possible that fooling is not necessary, or that perfect fooling can be achieved with simpler simulations. If fooling is not used, the potential problems that may arise and their solutions would fall within the traditional problem of AI boxing and containment (e.g., \cite{armstrongforthcoming,bostrom2014superintelligence}).

\textbf{Computational requirements. } The computational requirements for the simulations described here can be assessed by separately estimating the resources required for the agents and for the simulated environment. It is difficult to predict the requirements for a single AGI agent, but estimates have been suggested for the calculations per second required for a real-time functional simulation of the human brain. Such estimates range wildly from tens of Teraflops \cite{moravec2009rise} ($10^{13}$ FLOPS) to Exaflops and more ($10^{18}$ to $10^{25}$ FLOPS \cite{sandberg2008whole}). However, the highest estimates have been mostly suggested for detailed whole brain emulation approaches, which are unlikely to be the most computationally efficient approach to AGI, and may thus constitute an upper-bound on the actual requirements for computer-optimized implementations of the algorithms. A common intermediate estimate is for the required power to be of the order of tens of Petaflops ($10^{16}$ FLOPS) \cite{kurzweil2005singularity}, comparable to the performance of present day supercomputers.

As for the computational requirements for the simulated environment, multiple answers may be correct. Even today we are performing time-accelerated simulations in limited conditions, for example to train deep reinforcement learning agents, so there seems to be no strict lower bound on the required speed of the system. However, it is likely that more complex environments will be required in order to support AGI agents performing complex tasks, especially to allow progress in science and technology. While a certain degree of physically-detailed simulation of the real world may be required, a perfect simulation of the real world may not. Indeed, even present-day engineering software allow for part of the development in engineering to be performed in simulation (for example, using the COMSOL Multiphysics simulation software \cite{comsol2005comsol}).


Still, a perfect simulation may be required in case fooling of the agents was desired. For example, even if an imperfect simulation was sufficient to fool the agents, knowledge of the external world will be required to achieve progress in science and engineering, which could allow the agents to ultimately discover the truth. Nonetheless, while a perfect simulation would be computationally prohibitive with our current technology, we might be able to achieve it in the future \cite{baxter2001planetarium}. In any case, it seems unlikely that a perfect simulation will be required for the \emph{minimal simulation} and thus to start the Temporal Singularity.


\textbf{When. } If we assume that the agents require a computational power on the order of the average current estimates for the computational power of the human brain, and a linear scaling of the total requirements with the number of agents and temporal speedup, with negligible environment overhead, then the computational requirements for a minimal simulation of tens to hundreds of agents at faster than real-time may be as low as $10^{18}$ to $10^{21}$ FLOPS (e.g., $10^{16} \cdot 100 \rightarrow 100$ agents in real-time or $10$ agents at $10\times$ faster than real-time). If the Moore's law continues to hold, the world's most powerful supercomputer could achieve the required speed between the years 2020-2040, or alternatively individual home workstations between the years 2055-2075. Specialized hardware may however be developed to provide faster increases in the computational power in the future, as it has happened for example in the specific case of deep learning with the development of specialized accelerators like the Tensor Processing Unit (TPU) \cite{jouppi2017datacenter}. It is also interesting that these estimates are similar to current estimates for the development of AGI, which could be an important requirement for the simulations.

\textbf{Allocation of the resources. } We may further wonder how the available computing resources could be allocated between different processes to achieve the highest problem-solving capabilities of the system. For example, increased computation could be traded off between creating a larger number of agents, increasing the speed of the simulation and thus its temporal speedup compared to the external world, increasing the cognitive capabilities of the individual agents or simulating more complex environments. It may thus be required for the resources to be re-allocated dynamically depending on the state of technology.

\textbf{Potential limitations. } Even if the minimal simulation would be possible, there may be other limitations that could prevent or limit the Temporal Singularity. For example, it may be that temporal acceleration will not be the most efficient allocation of the computing resources, so that creating a larger number of agents or stronger ASIs will produce the best results. However, the fact that artificial agents working at faster than real-time are already being used and the potential advantages of simulating societies and civilizations suggest that this is unlikely to be the case. Temporal acceleration may also be helpful to speed up the solution of time-critical problems given a current level of intelligence of the available agents, in case improving the cognitive capabilities of the agent would prove difficult and more time-consuming. Finally, increasing the number of simulated agents may ultimately be limited by the intrinsic problem solving speed of each agent, which could be then trivially improved with temporal acceleration.

Another potential limit is that perfect simulations may be required to enable practical progress in science and technology. However, even present day research involves significant portions of time for theoretical work and simulations, so there seems to be a margin of speedup that can be achieved. Moreover, even if the first environments limited the potential to advance science and technology, it would be possible to iterate between time-accelerated work performed inside the simulation and prototyping, testing, and conducting experimental work in the external world, whose results and data could be fed back into the simulation to start the next cycle. Also, some type of theoretical work like in mathematics, computer science, philosophy and others may not need frequent access to data from the external world, suggesting that it should still be possible to benefit greatly from the temporal speedups of these simulations. In any case, interaction with the external world will always be required for maintenance and upgrades, to manufacture the newly developed technologies, and to acquire experimental data \cite{hutter2012can}. This dependency may ultimately limit the maximum speedups that can be achieved or their rate of growth. Still, even relatively low effective speedups could be highly beneficial. Further, the processes performed in the external world may be optimized inside the simulations to avoid wasting external time, for example by providing efficient instructions distributed among a large number of external world agents, although this may involve risks in the context of AI boxing \cite{armstrongforthcoming,bostrom2014superintelligence}.



\section{Implications}
\label{sec:implications}

\textbf{Advantages. } Similar to the Technological Singularity, the Temporal Singularity would produce a runaway increase the rate of growth of scientific and technological progress. In addition, however, it would also allow the study of the potential future of advanced intelligent civilizations and societal structures that will be required to be stable for extremely long intervals of time, which could be useful for scientific purposes and may provide invaluable information, thus impacting our society and guiding the future of our own civilization in a safe and beneficial way. In the extreme, we might be able to simulate civilizations with characteristics similar to our own, experimenting new societal designs and conditions. Finally, due to the potential for significant technological development, there is a clear competitive advantage for the first entity that will achieve a minimal simulation, even at moderate temporal speedups, whether it would be governments or private companies.

\textbf{Dangers. } In general, the Temporal Singularity shares all the potential dangers related to AGI/ASI and the Technological Singularity (see for example \cite{bostrom2014superintelligence}), and in particular to the problem AI boxing \cite{armstrongforthcoming}. However, the problems may be worse in this context, as even moderate temporal speedups would make it difficult to track the events inside the simulation. Finally, the same extreme progress in technology in a short span of time also constitutes a potential danger, as our society may not be capable of metabolizing it in the available time. For example, we can try to imagine what could have happened if we abruptly produced not just the technology, but a full stockpile of thermonuclear weapons during the Middle Ages.

\textbf{Fermi Paradox. } The Fermi Paradox is the contradiction between the apparent high likelihood of the existence of other intelligent civilizations in our galaxy or in the universe and the current lack of evidence of any. The Temporal Singularity leads to interesting implications in this context. First, if intelligent civilizations would achieve a degree of technology similar to our present one, and in particular develop computing systems, it may turn out to be almost inevitable that at some point they would produce a Temporal Singularity. Time-accelerated simulations could thus be part of some or all the possible intelligent civilizations, providing advantages like achieving a practical `subjective immortality' within the simulated environments, either for the individual agents or for their civilization as a whole, and subjectively delaying its demise due to the heat death of the universe or earlier extinction events. This can apply to either the external agents `moving into' the simulation, or for the simulated agents themselves as `mind children' progeny, as put by Hans Moravec \cite{moravec1988mind}, which could then possibly imply an abundance of post-biological civilizations in the universe \cite{dick2008postbiological,dick2003cultural}. An interesting possible outcome of this process is that time in the real world would be an important resource, and the speed of space colonization and communication, that is already considered slow, would become unbearable. Future civilizations may then prefer to avoid large-scale galactic colonization.


It is interesting to note that the Temporal Singularity shares features with the transcension hypothesis \cite{smart2012transcension} in the inevitable search for more energy and computing power, but ultimately produces opposite predictions, as in the transcension hypothesis advanced civilizations would try to slow down their subjective time by approaching black holes, rather than to accelerate it, in order to forward time travel to a time where all civilizations may ultimately meet and merge, and to optimize the acquisition of information.

On a negative side, the potential dangers that arise from this technology may constitute a `Great Filter' \cite{hanson1998great} that very few civilizations survive, thus explaining the Fermi paradox. However, time-accelerated simulations may also be used to escape traditional Great Filters by quickly providing us with solutions in time-critical situations, including for example impacts of asteroids detected with short notice or the Berserker scenario, in which an advanced intelligent civilization may attack any newly emerging civilization. 


Finally, a prediction of the Temporal Singularity in the context of the Fermi paradox can be made in the rapid increase in the power used by a civilization, tracking the super-exponential progress in technology, which could progress from using the resources of its host planet to those of its entire solar system within decades rather than millennia. Further, depending on the physical limits of technology, it may be possible that at least partial Dyson spheres \cite{dyson1960search} or Matrioska brains \cite{bradbury2001matrioshka} would be constructed in a relatively short time. The idea is particularly interesting as present day technology should be capable of detecting even partial neighboring Dyson spheres by changes in the infrared radiation of their host star \cite{wright2014g1,wright2014g2,griffith2015g3}. A prediction of the Temporal Singularity in the context of the Fermi Paradox is then on the speed of construction of such mega-structures.



\section{Conclusion}

We have explored the idea that progress in computing and artificial intelligence can lead to time-accelerated simulated civilizations unfolding in short time intervals in the external world, due to a runaway increase in the rate of growth of scientific and technological progress they could produce, that would quickly increase the temporal speedups of the simulations themselves, ultimately resulting in a `Temporal Singularity'. The potential advantages and dangers of such simulations have been briefly explored together with some implications of the Temporal Singularity on the Fermi paradox.

The medium-term relevance of the topic comes from the potentially relatively low computational power required to start the process, which could be as low as $10^{18}-10^{21}$ FLOPS and thus be available within the next decades, making the Temporal Singularity theoretically possible before the end of the century, and possibly in its first half.

As a final remark, it is interesting to note that given the great competitive advantages of running a simulation of the type described here, it is virtually inevitable that if it will ever be technically possible to create it, it will be created. It should be noted, however, that this is unlikely to happen in a discontinuous way, but rather we should expect an incremental progress, for example, starting from the simple advantage of temporal speedups in simulated environments for training artificial narrow intelligences (ANIs), as is already being done, to perhaps accelerating simulated `childhood' development and training of AGIs, to actual simulated multi-agent systems, building towards complete societies and civilizations following the increase in the available computing power.



\vspace{0.3in}
\noindent \textbf{Acknowledgments.} I would like to thank Ivana Kolorici and Renato Spigler for the helpful discussions and comments and the anonymous reviewers for the useful suggestions and references, from which this manuscript benefited significantly.




%
%
\bibliographystyle{splncs03}
\bibliography{./references.bib}  

\end{document}